\documentclass[review]{elsarticle}

\usepackage{amssymb}
\usepackage{graphicx}

\journal{Pattern Recognition Journal}
\usepackage{lineno,hyperref}
\usepackage{amsmath}
\usepackage{cleveref}[2012/02/15]

\crefformat{footnote}{#2\footnotemark[#1]#3}
\usepackage{booktabs}
\usepackage{bm}
\usepackage{amssymb}
\usepackage{multirow}
\DeclareMathOperator{\arccosh}{arcosh}
\usepackage{natbib}
\usepackage[export]{adjustbox}
\usepackage{soul}
\usepackage{xcolor}
\modulolinenumbers[5]
\usepackage{verbatim}
\usepackage{caption}
\usepackage{subcaption}
\usepackage[export]{adjustbox}
\usepackage{float}

\newcommand{\modelname} {COSKAD}
\newcommand{\hrub}{\emph {HR-UBnormal}}
\newcommand{\hrav}{\emph {HR-Avenue}}

\newcommand{\perfHRSTC} {77.1} % numbers

 \newcommand{\added}[1]{\textcolor{black}{#1}}
 \newcommand{\addedadded}[1]{\textcolor{black}{#1}}
\newcommand{\aadded}[1]{\textcolor{black}{#1}}

\usepackage{xcolor}
\usepackage{pifont}

\newcommand{\citen}[1]{#1 \textit{et al.}}
\newcommand*\colourcheck[1]{%
  \expandafter\newcommand\csname #1check\endcsname{\textcolor{ForestGreen}{\ding{52}}}%
}
\usepackage{colortbl}

\newcommand{\cmark}{\ding{51}}
\newcommand{\grayout}[1]{\textcolor{gray}{#1}}

\DeclareRobustCommand{\myparagraph}[1]{\noindent\textbf{#1}}

\begin{document}

\begin{frontmatter}

\title{\added{Contracting Skeletal Kinematics for Human-Related Video Anomaly Detection}}

\author[sap_cs]{Alessandro Flaborea\corref{mycorrespondingauthor}\fnref{equal}}

\cortext[mycorrespondingauthor]{Corresponding author}
\fntext[equal]{The first three authors contributed equally.}
\ead{flaborea@di.uniroma1.it}

\author[sap_cs]{Guido Maria D'Amely di Melendugno\fnref{equal}}
\author[sap_ds]{Stefano D'Arrigo\fnref{equal}}
\author[sap_ds]{Marco Aurelio Sterpa}
\author[sap_ds]{Alessio Sampieri}
\author[sap_cs]{Fabio Galasso}

\address[sap_cs]{Department of Computer Science, Sapienza University of Rome, Italy}
\address[sap_ds]{Department of Computer, Control and Management Engineering, Sapienza University of Rome, Italy}

\begin{abstract}

Detecting the anomaly of human behavior is paramount to timely recognizing endangering situations, such as street fights or elderly falls. However, anomaly detection is complex since anomalous events are rare and because it is an open set recognition task, i.e., what is anomalous at inference has not been observed at training. We propose \modelname, a novel model that encodes skeletal human motion by a graph convolutional network and learns to COntract SKeletal kinematic embeddings onto a latent hypersphere of minimum volume for \added{Video} Anomaly Detection. We propose three latent spaces: the commonly-adopted Euclidean and the novel spherical and hyperbolic. \added{All variants outperform the state-of-the-art on the most recent \emph{UBnormal} dataset, for which we contribute a \emph{human-related} version with annotated skeletons. COSKAD sets a new state-of-the-art on the human-related versions of \emph{ShanghaiTech Campus} and \emph{CUHK Avenue}, with performance comparable to video-based methods.} Source code and dataset will be released upon acceptance.
\end{abstract}
\label{sec:abstract}

\begin{keyword}
anomaly detection \sep open set recognition \sep hyperbolic geometry \sep kinematic skeleton \sep graph convolutional networks 
\end{keyword}

\end{frontmatter}

% \linenumbers

\section{Introduction}

Anomaly Detection (AD) is a broad and well-studied field in computer vision, which aims to detect events that deviate from normality automatically~\aadded{\cite{CEVIKALP2023109385}}. More precisely, the task is to detect anomalous events in footage and label the corresponding frames as abnormal. AD is a complex and multifaceted field with applications beyond just video surveillance~\cite{sultani18}. Many techniques have been successfully applied in several real-world scenarios, including monitoring elderly individuals~\cite{PRENKAJ2023102454, Flaborea_2023_CVPR}, industrial systems~\cite{Flaborea_2024_CVPR}
% \cite{atha2018evaluation}, 
and \aadded{point clouds~\cite{CAO24}}.
% social networks~\cite{liu2015social}.

While significant progress has been made in AD in recent years, this task still presents several challenges: (1) anomalous events are rare in real-world scenarios, and this is reflected in the imbalanced distribution of normal and anomalous events in public AD datasets. (2) Anomalous events are challenging to identify since abnormal actions can involve either frenetic movements (e.g., fights) or very still postures (e.g., faints), resulting in a context-dependent definition of anomaly, which varies among public datasets. Furthermore, the same anomalous action can vary significantly among individuals, adding to the intra-class variation that must be considered. Finally, (3) when working with videos of humans, it is crucial to consider privacy and fairness concerns, such as avoiding violations of individuals' rights or exploiting social biases. 

\begin{figure}[!t]
    \begin{center}
\includegraphics[width=0.9\linewidth]{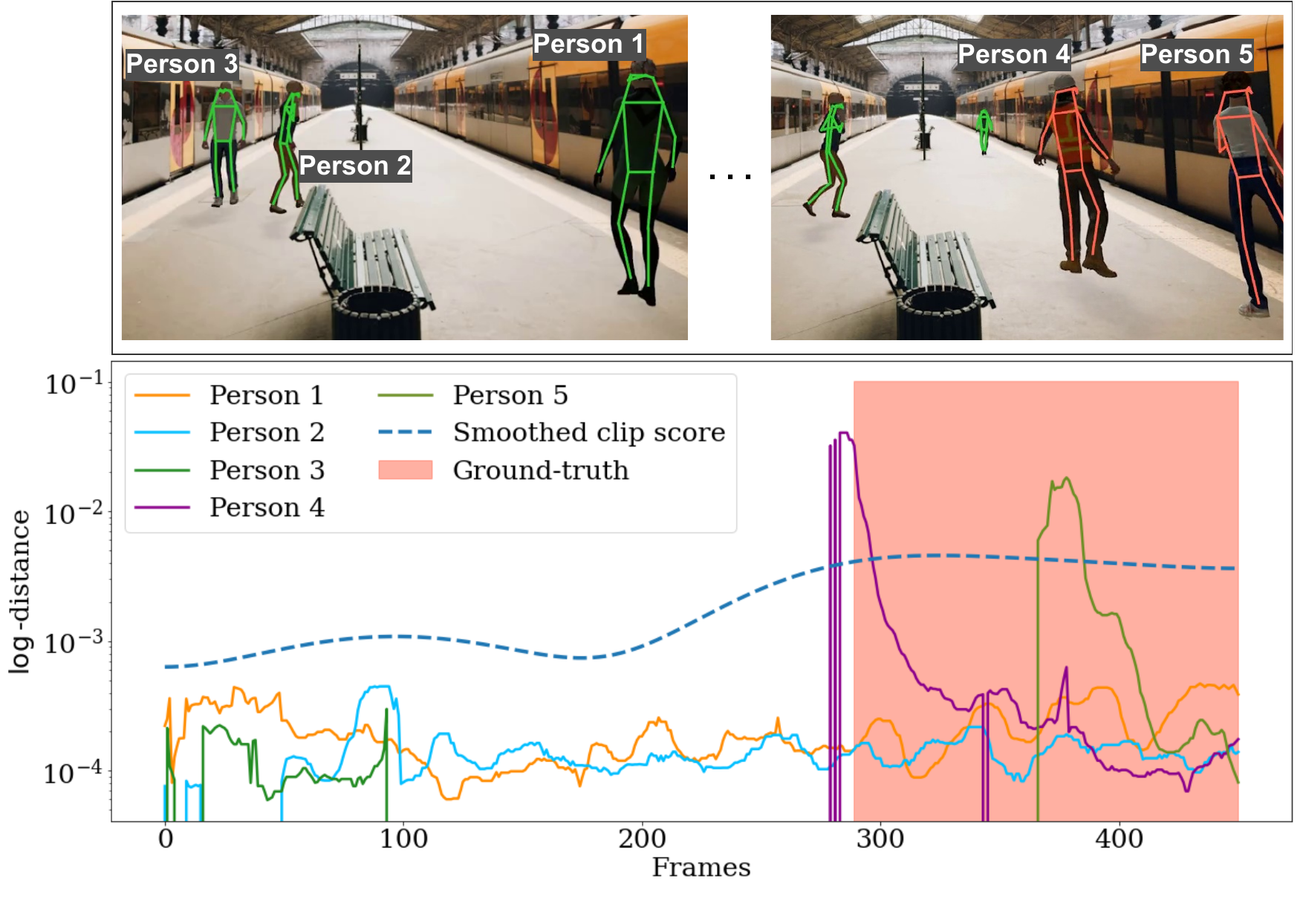}
    \end{center}   
    \caption{
    Anomaly score provided by \modelname\ on a clip from the UBnormal dataset. \modelname\ correctly classifies the motion of the two staggering characters (red skeletons in the upper-right picture) in the last part of the clip as anomalous.
    }
    \label{fig:teaser}
\end{figure}

In this work, we propose \modelname. This novel end-to-end model infers abnormality COntracting SKeletal (\modelname) embeddings in the latent space and opens up a novel investigation of the latent metric space encoding the regular motions. The proposed method tackles all the limitations mentioned above. First, it adheres to the \textit{One-Class-Classification} (OCC)~\aadded{\cite{morais19,markovitz20,sun24}}
%%%% removed hasan16, liu18
% \cite{hasan16, liu18, luo19, morais19, markovitz20} 
protocol, which simulates the scarcity of abnormal events exposed in (1).  
Moreover, \modelname\ exploits the compact spatio-temporal skeletal representation (cf. Fig. \ref{fig:teaser}) of human motion instead of raw frames, typically employed with video-based methods. This modality comes with several advantages; 
first, recent work~\cite{xu22stars} in the motion-related fields of Pose Forecasting and Action Recognition has proven the superiority of this modality with respect to the raw video frames for motion representation. Exploiting this input modality, our proposed model is more robust to slight motion deformations that represent the intra-class variability described in (2) while preserving the ability to distinguish between different actions, e.g., walking vs. running, which is crucial in AD. Furthermore, modeling agents as kinematic graphs allows for separating the person detection task (handled as a preprocessing step) from the anomaly detection task, resulting in a more computationally efficient and context-agnostic method. Although skeletal representations are more compact than raw video frames, they contain the necessary information for effectively characterizing motions, as demonstrated by the state-of-the-art results of skeleton-based methods~\cite{luo19,morais19,markovitz20}. Finally, the skeletal representation is more privacy-preserving (3) since it ignores biometric details, representing all samples with anonymous tensors of coordinates.
Consistently with this setting, we adopt the Human Related (HR)\cite{morais19} split of the datasets, which corresponds to a version of the dataset that admits only anomalies generated by humans, e.g., people fighting, disregarding anomalies coming from the context scene, e.g., a car proceeding on the pavement.

The proposed model comprises two components: an encoder and a projector module. \addedadded{Regarding the first}, differently from previous related works, which use either ST-GCN~\cite{markovitz20,yan18,luo17} or a Recurrent Neural Network~\cite{morais19} to encode the human motion, \modelname\ implements its encoder as a Space-Time-Separable Graph Convolutional Network~\cite{sofianos21}. 
As far as we know, this is the first study to adapt it for skeletal-based AD and to expose a comparison among different GCN-based encoders (presented in Sec.~\ref{encoderablation}).
\addedadded{The second model component is the projector, for which we draw inspiration from} SSL, where it has been shown to play a crucial role\aadded{~\cite{simclr,byol}}, which we confirm here in detailed ablation studies (cf. Sec.~\ref{projectorablation}). 

\addedadded{We propose to learn COSKAD by a data-driven metric objective in the latent space that minimizes the distance between the skeletal embeddings and a center.} We call the distance minimization \emph{contraction}, as it forces normality to concentrate around an origin.
Dealing with a metric objective, we further propose a novel investigation of how the latent distribution might be \textit{altered}, \addedadded{\textit{expanded}, or \textit{condensed}} using the peculiar metric properties of three \addedadded{corresponding} manifolds: the Euclidean~($\mathbb{R}^n$), the hyperbolic~($\mathbb{H}^n$), and the n-Sphere~($\mathbb{S}^n$).
\addedadded{Euclidean spaces are default choices for representation and metric learning. In our setup, the distance of a sample from the center is proportional to the likelihood of the sample being anomalous. Forcing normal samples to lie in a narrow region of the Euclidean space has been adopted by previous literature~\cite{ruff18}.}
\addedadded{By contrast, the hyperbolic space is characterized by exponential volume growth at increasing radii, which affects contraction by accommodating larger-scale variations in the normality region. Third, we experiment with the spherical space, where all the points are condensed in a compact region as they lie on the sphere's surface, and the distance between them is bounded above by the length of half orthodrome.} 
\addedadded{To the best of our knowledge}, this is the first work to have studied different latent spaces for skeleton-based anomaly detection. 
The proposed model is simple, lightweight, and effective, as we illustrate in extensive experiments, where \modelname\ outperforms state-of-the-art (SoA) models (including some video-based techniques) on three challenging benchmarks: \emph{HR-ShanghaiTech Campus}~\cite{markovitz20,luo17}, \hrav~\cite{morais19,lu13}, and the recent \emph{UBnormal}~\cite{acsintoae22}.

\addedadded{We perform} an ablative examination of several of our model's key features. Beyond a thorough ablation study on the main modules of our proposed \modelname, we also compare the encoder-based architecture proposed with an autoencoder and compare two alternative strategies to define the hypersphere center, extending the seminal study of \cite{ruff18}. 
Finally, we propose a novel HR version of UBnormal~\cite{acsintoae22}, dubbed \textit{HR-UBnormal} as an additional contribution. We create HR-UBnormal by filtering out anomalous events that do not involve human individuals, e.g., we remove scenes of fire and car accidents unless people are involved.
We leverage an established Pose Estimator \cite{fang2017rmpe} and refine its results with a Pose Tracker \cite{xiu2018poseflow} to extract human poses in each UBnormal's video frame. So, HR-UBnormal contains human-related anomalies and human skeletons at all frames, inspected for accuracy and temporal consistency.
To summarize, the contribution of this paper is threefold:
\begin{itemize}
    \item We introduce \modelname, a simple, end-to-end, and effective model that surpasses SoA results on three public benchmarks.
    \item We conduct an in-depth study on three different manifolds as latent spaces, explore their intrinsic properties and thoroughly analyze their effects on our novel AD system.
    \item We introduce a new \textit{HR} version of UBnormal with a filtered selection of clips featuring human-related events and an extended set of human body pose annotations.
\end{itemize}
\label{sec:introduction}

\section{Related work}

Anomaly Detection (AD) is a multi-faceted field with applications in several domains (see~\aadded{\cite{bogdoll22}).}
%%%%%% removed hilal22 %%%%%%%%
% \cite{bogdoll22,hilal22} for surveys). 
This work focuses on skeleton-based anomaly detection, a type of video-based AD that involves analyzing the movements and poses of human bodies in a video.
In this section, we compare works that most closely relate to ours, distinguishing error-based and score-based video AD techniques and skeleton-based models. 

\subsection{Video AD} 
Early proposed methods analyze the trajectories of agents in the video to unearth those that differ from normality.
% ~\cite{Jiang11}.
%%%%%% removed calderara11, Jiang11
% \cite{Jiang11, Calderara11}. 
More recent deep learning methods for Video AD can be roughly collected into two categories: error-based or score-based.

\myparagraph{Error-based methods.} These methods attempt to detect anomalies through a generative process in which a model produces new video frames, which are then compared with ground truth. These methods assume that a model trained with only normal data will struggle to generate the anomalous frames, producing a more significant error that can be directly used as the anomaly score. 
%%%%%%% removed hasan16
% \cite{hasan16} used convolutional AutoEncoder (AE) and defined the input volume as a stack of sequential grayscale frames. 
\citen{Gong}~\cite{gong19} during the training phase builds a memory of the most representative normal poses. During the inference phase, for each sample, it finds the most similar example in memory and estimates its ground truth distance. 
\aadded{Several works rely on a hybrid objective that defines the anomaly score of each frame as the sum of the reconstruction error and the prediction error. This strategy has recently been adopted by \citen{Tang}~\cite{TANG20}, defining a model based on GAN for frame prediction and reconstruction, and then refined by \citen{Sun}~\cite{sun24}, proposing to adopt two separated generators for the two tasks. \citen{Qiu}~\cite{qiu24} defined a U-Net-shaped network tasked to predict and reconstruct the video frames. Within this model, the authors defined a two-scale deep clustering module that helps sharpen the differences between normal and anomalous samples and suppress the redundant information in qualitatively similar consecutive frames.}
Unlike these methods, \modelname~relies on a GCN-encoder, an ideal tool for exploring relationships between body joints over time in the kinematic graph and extracting semantically consistent latent embeddings.

\myparagraph{Score-based methods.} These approaches have been extensively studied~\aadded{\cite{ruff18,wang21}.}
%%%%%%% removed tax04
% \cite{ruff18,tax04,wang21}. 
They derive abnormality in videos by monitoring some quantity extracted from the embeddings produced by the deep network. For example, \citen{Sabokrou}~\cite{sabokrou17} proposed a two-stage method in which videos are divided into cubic patches and first analyzed with Gaussian classifiers to exclude the least relevant patches, e.g., background. Then, the remaining candidates are processed by a more complex CNN. 
In contrast, \modelname~looks at the latent space positions occupied by input embeddings to derive clues of abnormality. While working with images rather than video, Deep Support Vector Data Description (DSVDD)~\cite{ruff18} and OC4Seq~\cite{wang21} are two methods related to \modelname, as both employ a DSVDD objective seeking to minimize a sphere enclosing the generated embeddings. While we also employ an SVDD objective, our model learns in an end-to-end way to map the representation of the normal samples in the latent space while encoding semantic representations thanks to its separable GCN encoder. As far as we know, this work is the first to propose the SVDD objective for Video AD with skeleton-based representations.

\subsection{Skeleton-based AD}
\citen{Morais}~\cite{morais19} first introduce the skeleton-based representation in AD; their method presents a two-branches architecture for reconstruction and forecasting modeled as GRUs. 
\citen{Luo}~\cite{luo19} set up the problem as motion forecasting and define their model by stacking layers of ST-GCN~\cite{yan18} followed by an MLP forecasting module. While they introduce the use of GCN, the adjacency matrix is fixed and does not allow the exploration of intra-frame and intra-joint relationships, improving spatio-temporal features encoding~\cite{sofianos21}. \citen{Markovitz}~\cite{markovitz20} propose a two-stage network in which they train an autoencoder (built on ST-GCN~\cite{yan18}), and, in the second stage, it clusters the produced embeddings in the latent space.
These clusters should represent the normality styles in the train set, but it is challenging to spot the optimal number of clusters. Differently, \modelname~has an end-to-end approach and solves the problem of the number of clusters by forcing the entire train set into the same latent region, constraining the distances to a common center.
\label{sec:relworks}

\section{Methodology}

In this section, we describe our proposed model focusing on its modules and the steps taken to train and assess it.

\begin{figure}[!t]
    \centering
    \includegraphics[width=\linewidth]{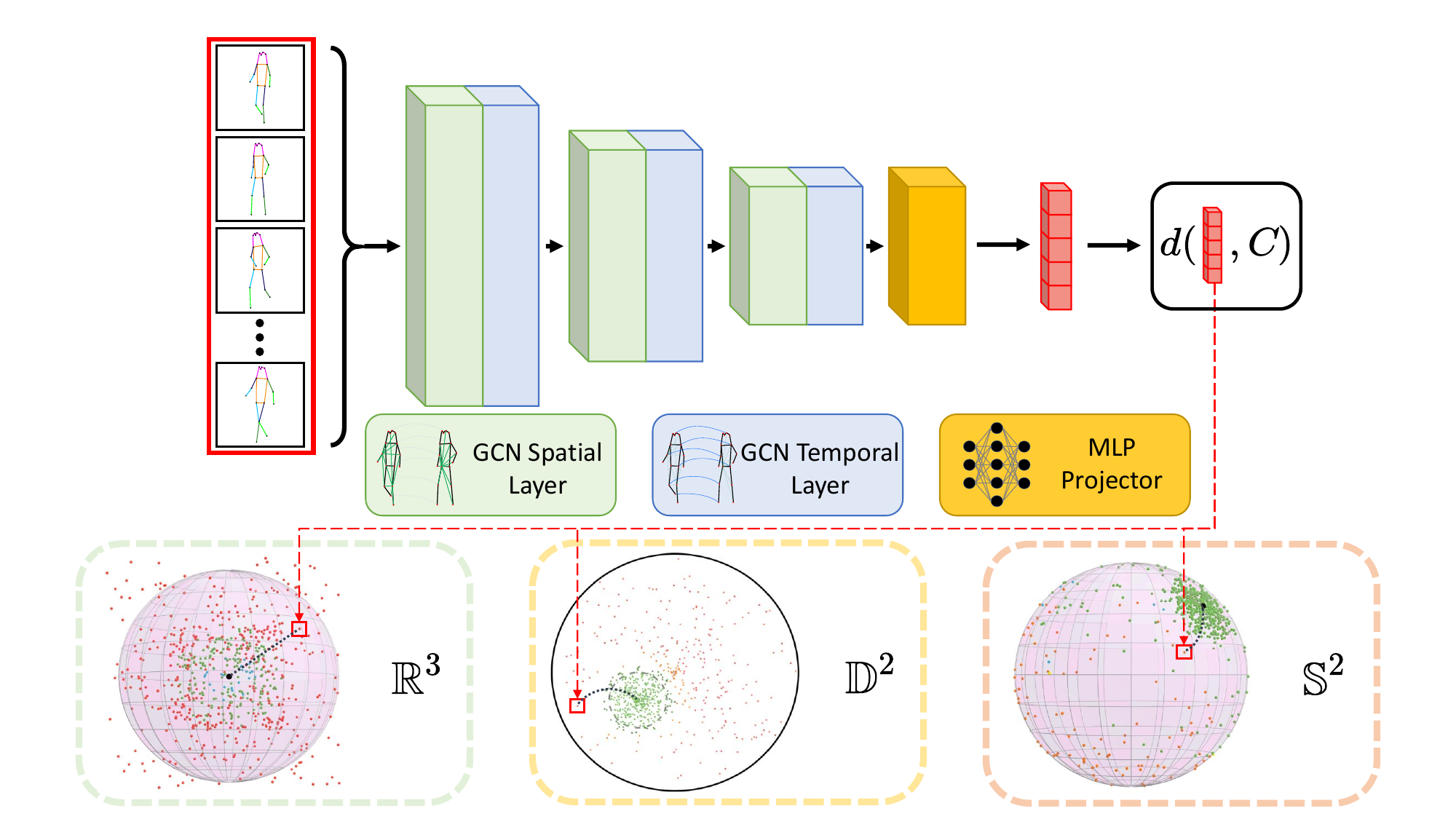}
    \caption{The overall architecture of \modelname. The model combines an STS-GCN-based~\cite{sofianos21} encoder (\added{light} green \added{and light blue} blocks) with a projector module (yellow block)
    After projection, the latent representation (\added{red} vector in the figure) is embedded into the latent space.
    We propose and evaluate 3 variants of the latent space: \textit{Euclidean} $\mathbb{R}^n$, \textit{spherical} $\mathbb{S}^n$, and the \textit{hyperbolic} modeled with the Poincaré Ball $\mathbb{D}^n$. 
    \added{During training, the embeddings are constrained to accumulate in a narrow region in the chosen manifold by reducing the distance between the motion embedding and the common center.} The sequences mapped further from the center are interpreted as anomalous during inference.
    }
    \label{fig:pipeline}
\end{figure}

We assume the human body kinematics to be available as skeleton representations for a few given frames (cf. Sec. \ref{old_datasets} for the details on the skeleton sequence extractions for the proposed HR-UBnormal dataset).
These spatio-temporal graph inputs are fed to \modelname~which, as illustrated in Fig.~\ref{fig:pipeline}, relies on two key components: a separable graph encoder and a projection module. The encoder processes the input graph and produces embeddings representing the motion of each individual. The projector adapts the embedding provided by the encoder for the mapping in the latent space. Both modules are jointly trained with a spatial minimization objective, which aims to catch the correspondences among samples belonging to the same class. Finally, we define a novel metrical objective in the latent space in order to guide the training and consider three different manifolds as latent spaces: the \textit{Euclidean Space} (\added{$\mathbb{R}^n$}), the \textit{Poincaré Ball} ($\mathbb{D}^n$), and the \textit{n-Sphere} ($\mathbb{S}^n$), to inherit their specific metric properties.

\myparagraph{Formulation.} The motion trajectories consist of $V$ joints per actor in each frame tracked across all the frames ($T_{actor}$) where the actor is present.
We apply a temporal sliding window crop on the trajectories to get sequences of $V$ joints' spatial positions for $T$ adjacent time frames. Finally, we organize the input signal as a graph $\mathcal{G}=(\mathcal{V},\mathcal{E})$, with $TV$
nodes $x_i \in \mathbb{R}^{C}$, where $C=2$ stands for the $x,y$ joint coordinates, and with edges  $(i, j) \in \mathcal{E}$, represented by a spatio-temporal adjacency matrix $A^{st}\in \mathbb{R}^{VT \times VT}$, relating all joints to all others across all observed time frames.

\subsection{Encoder \added{and} Projector module} \label{encodermethod}

We encode the body kinematics of a person with a SoA variant~\cite{sofianos21} of Graph Convolutional Network (GCN)~\cite{kipf16}. GCNs are the go-to choice in the kinematic-related fields of Pose Forecasting \cite{xu22stars,sampieri_sesgcn,Rahman_2023_CVPR} and action recognition \cite{yan18,song22}.
\modelname\ leverages a separable GCN-encoder designed to capture spatio-temporal signals and produce consistent features. 
Specifically, the encoding performs a factorization of the adjacency matrix $A$ into two learnable submatrices ($A_s,\ A_t$) responsible for spatial and temporal interconnections, respectively. $A_s \in \mathbb{R}^{T \times V\times V}$, the \emph{spatial adjacency matrix}, learns the relationships between different joints in each frame by learning their interdependence. In contrast, $A_t\in \mathbb{R}^{V\times T \times T}$, the \emph{temporal adjacency matrix}, deals with the connections between different temporal instants for each joint. 

This strategy, also adopted by \cite{sofianos21} in the context of pose forecasting, ensures effective encoding of the spatio-temporal features of the input graph by learning and exploiting the joint-joint and time-time relationships that characterize human motion. In addition, this factorization results in a considerable reduction in the number of parameters ($\sim$ 4x with respect to a plain GCN) since it does not consider the relationships between different joints in different frames. 

We formulate a single encoder layer as:
\begin{linenomath}\begin{equation}
    X^{l+1} = \sigma(A_s A_t X^l W)
    \label{eq:stsgcn}
\end{equation}\end{linenomath}
where $X^l$ is the input from the previous layer, $W \in \mathbb{R}^{C \times C'}$ are learnable weights and $\sigma$ an activation function. We stack four separable GCN layers interleaved with residual connections to encode the entire input sequence.
Overall, the results of our ablation studies (cf. Sec. \ref{encoderablation}) suggest that the separable GCN is the best encoder among the GCN architectures we evaluated. 

Reminiscent of recent works in Self-Supervised Learning (SSL) \cite{simclr,byol}, we explicitly define a projector module to refine the Encoder representation and accommodate it in the latent space. The projector comprises two identical blocks that iteratively process their input with a Fully Connected Module followed by a ReLU non-linearity and a Batch Normalization 
%%%%%%%%% removed ioffe (BN) - ablation
% \cite{ioffe15} 
layer. Despite its simplicity, we found it beneficial to add this module, as shown in Sec. \ref{projectorablation}.

\subsection{Objective} \label{projmethod}

The OCC formulation requires that the train sets contain elements of a single class.
Therefore, the design of the objective function is crucial, as it can only exploit the similarities that normal samples exhibit.
Many existing methods~\cite{luo19,morais19,gong19} 
%%%%%%%%%%% removed liu18 
% \cite{liu18,luo19,morais19,gong19} 
formulate the anomaly score using a proxy task, such as the reconstruction error. In Sec. \ref{autoencoderablation}, we experimentally confirm that this is suboptimal, as maintained in \cite{georgescu21_1}, since it does not align directly with the inference AD objective.

In this work, we define a metric objective that encourages the model to arrange the train samples in a narrow region in the latent space: inspired by \cite{ruff18}, the objective is defined as:
\begin{linenomath}\begin{equation}
    \min_\mathcal{W} \frac{1}{N}\sum^N_{i=1} d(\Phi(x_i, \mathcal{W}),\ c) + \alpha f(\mathcal{W}) 
    \label{eq:svdd}
\end{equation}\end{linenomath}
Where $\Phi$ represents the mapping in the latent space defined by \modelname, 
$\mathcal{W}$ \added{is} the set of its parameters, $x_i \in \mathbb{R}^{T \times V \times C}$ is an input sample, $d$ is a metric defined in the latent space, $f$ is a weight decay regularizing function, and $c$ is the center of the hypersphere. 
We follow \cite{ruff18} and initialize $c$ by taking the average of the final embeddings after an initial forward pass. Then, \cite{ruff18} suggests training the model with the parameter $c$ fixed. We argue that this practice may harm the training, leading to a non-optimal solution since the center is precalculated and its position does not evolve with the learning.

Differently from \cite{ruff18}, we introduce a novel and more tailored data-driven dynamic for the center: at the beginning of each training epoch, the position of the center is refined to be the centroid of the data's projection in the latent space. The benefit of this moving center is twofold. First, a data-driven center relieves the encoder and the projector learning not being constrained to accumulate projection around a fixed point. Secondly, it encourages \modelname~to explore the latent space to find a region that accommodates the representations, which we show to be beneficial in Sec. \ref{centerablation}, especially when the hyperbolic manifold is set to be the latent space.

\begin{figure}[t]
    \centering
    \centering
    \begin{subfigure}{0.25\textwidth}
        \includegraphics[width=\textwidth]{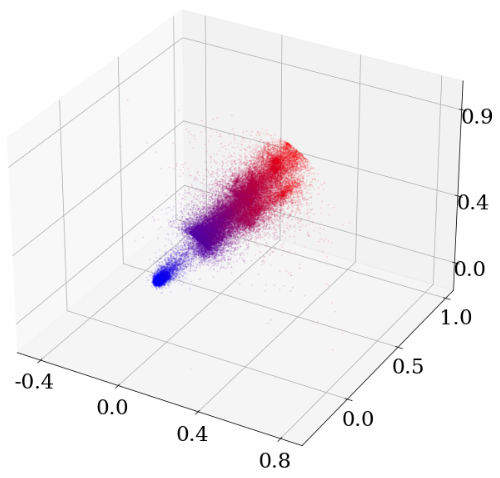}
        \caption{}
    \end{subfigure}\hfill
    \begin{subfigure}{0.25\textwidth}
        \includegraphics[width=\textwidth]{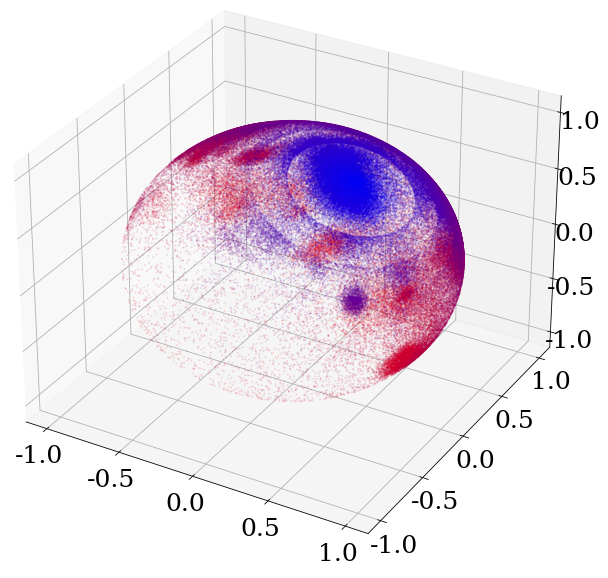}
        \caption{}
    \end{subfigure}\hfill
    \begin{subfigure}{0.27\textwidth}
        \includegraphics[width=\textwidth]{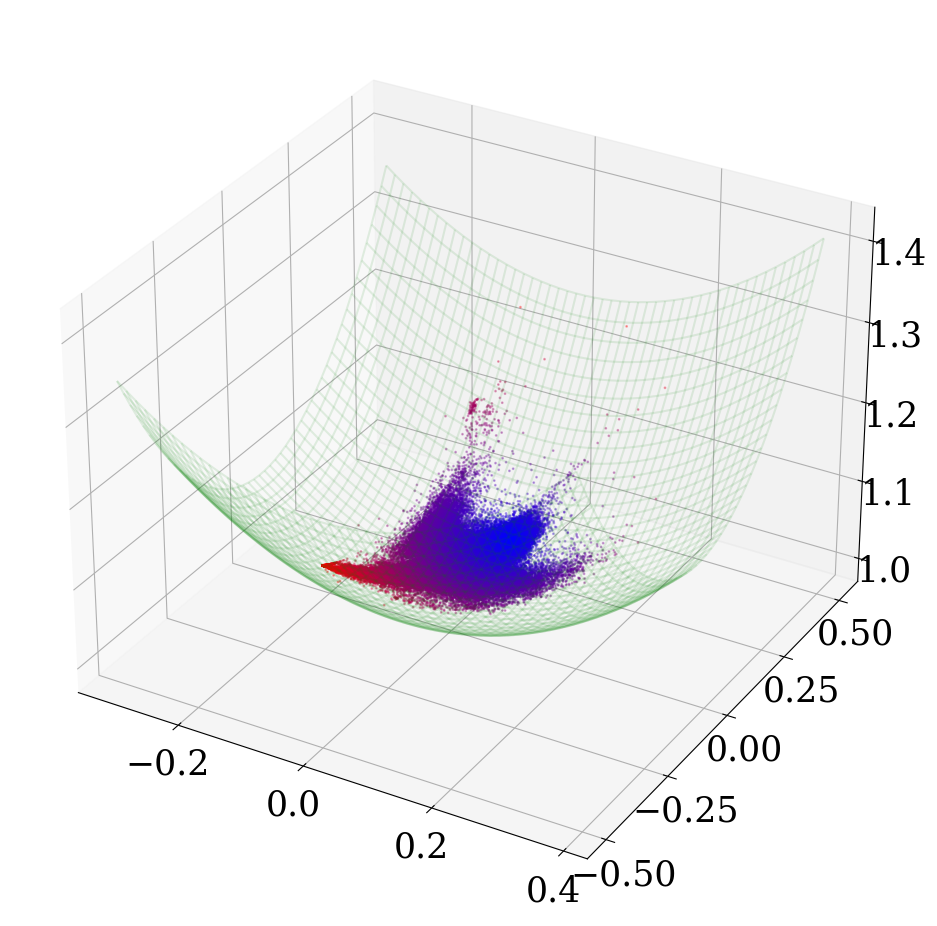}
        \caption{}
    \end{subfigure}

    \caption{Visualization of the UBnormal test set's latent vectors embedded in three different manifolds: (a) Euclidean, (b) spherical, and (c) hyperbolic. We retain the three dimensions with the highest variance and color-code the points according to their distance from the center, from blue (closest) to red (furthest).
    Distance is intended as the $L^2$ norm in the Euclidean case, the \textit{cosine distance} on $\mathbb{S}^n$, and the \textit{Poincaré distance} for the hyperbolic embeddings. In the hyperbolic case, we highlight in green the hyperboloid onto which the embeddings are projected for better visualization.}
\label{fig:spaces}
\end{figure}

\subsection{Latent Spaces} \label{latentmethod}
We propose to consider three diverse manifolds for embedding the input sequences. In fact, the objective of Eq. \ref{eq:svdd} forces the model to focus on features corresponding to common characters in the embeddings extracted from the encoder and to iteratively reduce their distances from a center. As a result, the method relies on a metric which, in turn, depends on the metric space $\mathcal{L}$ chosen as the latent space.
Motivated by this, we are the first to define the same objective on three manifolds, each with its own specific metric: the \textit{Euclidean space} $\mathbb{R}^n$, the \textit{spherical space} $\mathbb{S}^n\subset \mathbb{R}^{n+1}$, and the \textit{hyperbolic space} $\mathbb{H}^n\subset\mathbb{R}^{n+1}$. These spaces have a different curvature that causes the distance to behave differently in each manifold.

\subsubsection{Euclidean Latent Space} \label{eucl_met}
When $\mathcal{L}=\mathbb{R}^n$, the distance coincides with the $L^2$ metric $d_E(x, y)={||x-y||}_2$, and the model is defined substituting the distance $d$ with $d_E$ in Eq. \ref{eq:svdd}.

\subsubsection{n-Sphere Latent Space} \label{sphere_met}
With $\mathcal{L}=\mathbb{S}^n$, we modify the proposed \modelname~model, building on top of the \textit{S-VAE} presented in \cite{s-vae18}. We dub the model \textit{\modelname-radial} since it constrains inputs onto the spherical surface with a fixed radius $\mathbb{S}^{n} = \{x \in \mathbb{R}^{n+1}:$ $\|x\|_2 = 1\}$; the posterior of the normal data approximates a Power Spherical distribution $q_X(x; \mu, \kappa)$ \cite{decao2020power}, with a constraint $L_{dir} = \frac{1}{N} \sum_{i=1}^N (1-x\cdot c)$ to push the samples close to the empirical mean direction $c$. The target loss is $L = \gamma L_{rec} + \phi L_{dir} + \beta L_{KL}$, where $L_{rec}$ defines the objective for the Variational AutoEncoder reconstruction, $L_{KL}$ is the Kullback-Leibler divergence, $\gamma, \phi, \beta \in \mathbb{R}$ hyperparameters. The anomaly score is solely given by the \textit{cosine distance} between any sample $x$ on  $\mathbb{S}^{n}$ and the empirical mean $c$ computed at train time.

\subsubsection{Hyperbolic Latent Space} \label{hyp_met}
When $\mathcal{L}=\mathbb{H}^n$, we model it with the \textit{Poincaré Ball} which coincides with the unit euclidean open ball $\mathbb{D}^n$ endowed with the \added{Riemmanian metric $$g_x^\mathbb{D} = \frac{2I^n}{1-{||x||}^2},$$ where $I^n$ represents the Identity matrix. This metric tensor induces the} distance

\begin{linenomath}\begin{equation}
	{d_{\textcolor{black}{\mathbb{D}}}} (x, y) = \arccosh \bigg(1+\frac{2 ||x-y||^2}{(1-||x||^2)(1-||y||^2)}\bigg).
	\label{eq:hyp_dist}
\end{equation}\end{linenomath}

The distances between the points on this manifold increase exponentially with distance from the origin $O$. 
 To take full advantage of this property, we let the cluster center move in the Poincaré Ball in a data-driven fashion, but we also experienced defining the center as a fixed point, as detailed in Sec.~\ref{centerablation}.
In Fig.~\ref{fig:spaces}, comparing the Euclidean (a) with the hyperbolic (c) case, the effect of this distance is shown: in the Euclidean case, farther points tend to be more spread since the hyperbolic distance $d_\mathbb{\added{D}}$ ensures a stronger attraction towards the center, i.e. $d_{\textcolor{black}{\mathbb{D}}}$ grows exponentially with the distance from the center. In this case, the model is obtained from \modelname~by adding a projection layer $\exp_0:\mathbb{R}^{n+1}\rightarrow \mathbb{D}^n$, which performs the exponential mapping to get the hyperbolic representation in $\mathbb{D}^n$. Thus, the objective defined in Eq.~\ref{eq:svdd} becomes:

\begin{linenomath}\begin{equation}
    \min_\mathcal{W} \frac{1}{N}\sum^N_{i=1} {d_{\textcolor{black}{\mathbb{D}}}} (\exp_0(\Phi(x_i, \mathcal{W})),\ c) + \alpha f(\mathcal{W}). 
    \label{eq:svdd_hyp}
\end{equation}\end{linenomath}

\subsection{Anomaly Score} \label{scoring}

At inference time, the anomaly score $s$ for each sample $x$ (representing a single agent in a time window composed by $T$ frames $\{f_1, f_2, ..., f_T\}$) is defined as the distance of its \modelname\ embedding from the center $c$:
\begin{linenomath}\begin{equation}
	s(x) = d(\Phi(x, \mathcal{W}), c).
	\label{eq:windowscore_d}
\end{equation}\end{linenomath}
To score a single frame $\bar{f}$, for each agent $p$, we first collect all the windows involving $p$ in a set $w^p$, and we restrict this set picking only the windows containing $\bar{f}$ to a set $w^p_{\bar{f}}$. Then, we calculate the score through Eq.~\ref{eq:windowscore_d} and thus set the score for $p$ at frame $\bar{f}$ to be the mean of those scores:
\begin{linenomath}\begin{equation}
	s^p(\bar{f}) = \frac{1}{|w^p_{\bar{f}}|}\sum_{x \in w^p_{\bar{f}}} s(x).
	\label{eq:persframescore}
\end{equation}\end{linenomath}
Finally, we get the score for a single frame $\bar{f}$ by a max pooling operation over all the agents present in the scene at frame $\bar{f}$:
\begin{linenomath}\begin{equation}
	s(\bar{f}) = \max_{p \in P_{\bar{f}}} \{s^p(\bar{f})\},
	\label{eq:framescore}
\end{equation}\end{linenomath}
where $P_{\bar{f}}$ is the collection of all the people present in the clip at frame $\bar{f}$.
\label{sec:methodology}

\section{Experiments}

\added{In this section, we evaluate the performance of \modelname\ on the human-related versions of three established benchmarks against state-of-the-art skeleton-based methods. Next, we discuss how our approach relates to techniques that additionally employ appearance.} The section is organized as follows: first, we illustrate \added{the datasets and the metric (Sec.~\ref{datasetsExp}).} Then, we describe the \added{results of the experiments and discuss the performance~(Sec.~\ref{comparison}).
The last part of this section provides details about the implementation and data preprocessing (Sec.~\ref{exp_setup}).} 
\begin{table}[t]
\centering
\renewcommand{\arraystretch}{1.3}
\resizebox{\columnwidth}{!}{
\begin{tabular}{cc|ccc|cc}
    \hline
    \multirow{2}{*}{\textbf{Dataset}} & \multicolumn{6}{c}{\textbf{\# frames}} \\
    \cline{2-7}
    & \textbf{\textit{Total}} & \textbf{\textit{Training}} & \textbf{\textit{Validation}} & \textbf{\textit{Test}} & \textbf{\textit{Normal}} & \textbf{\textit{Abnormal}}\\
    \hline
    \hline
    Avenue \cite{lu13} & 43,499 & 28,175 & - & 15,324 & 25,891 & 17,608 \\
        
    STC \cite{luo17} & 317,398 & 274,515 & - & 42,883 & 300,308 & 17,090\\
    UBnormal \cite{acsintoae22} & 236,902 & 116,087 & 28,175 & 92,640 & 147,887 & 89,015 \\
    \hline
    \hline
    HR-Avenue \cite{morais19}  & 42,883 & 28,175 & - & 14,708 & 25,891 & 16,992 \\
    HR-STC \cite{morais19} & 313,212 & 274,515 & - & 38,697 & 297,090 & 16,122\\
    HR-UBnormal (\textit{Proposed}) & 234,751 & 116,087 & 28,175 & 90,489 & 147,887 & 86,864 \\

\hline

\end{tabular}}
\caption{Overview of the three datasets chosen, CUHK Avenue, ShanghaiTech Campus, and UBnormal, as well as their human-related versions. 
}
\label{tab:datasets}
\end{table}

\subsection{Benchmarks} \label{datasetsExp}

The following sections describe the datasets and \added{introduce the metric we use to score anomalous frames in a video sequence.}

\subsubsection{Datasets} \label{old_datasets}

\myparagraph{UBnormal.} \added{UBnormal~\cite{acsintoae22} is the largest and most recent dataset for frame-level video anomaly detection, the only one to provide \textit{train-validation-test} splits that adhere to the \textit{Open Set} protocol, i.e., the sets of anomalous events for train, validation, and test are disjoint. It consists of 19 clips, which encompass both normal and abnormal events, for each of the 29 diverse background scenes. UBnormal has several distinctive features. First, as illustrated in Table~\ref{tab:datasets}, it contains more anomalous frames than those presented in CUHK Avenue or ShanghaiTech Campus. Additionally, it has labels for each anomaly at both the frame and pixel level, with annotations describing the anomalous actions (e.g., jumping, sleeping, stealing). UBnormal exhibits a larger list of anomalous event types than previous Video Anomaly Detection datasets: it includes 22 categories of anomalies, in contrast to ShanghaiTech Campus and CUHK Avenue which only encompass 11 and 5, respectively. Additionally, it introduces a variety of objects, such as cars and bicycles, to both the train and the test sets to avoid the anomaly being detected because the object was not seen during the training phase. In contrast to other benchmarks, it is synthetic and its virtual 3D scenes are created with Cinema4D with heterogeneous 2D backgrounds (e.g., streets, train stations, and office rooms, with different viewpoints and lighting conditions).
Since \modelname\ is trained in the OCC framework, we extract only the normal sample poses from the training set. Conversely, we maintain the original validation and test split in this setting.} 

\myparagraph{HR-UBnormal (\textit{Proposed}).}
We propose HR-UBnormal as an extension of the original UBnormal dataset with kinematic motion representations and a selected set of anomalies that relate only to human behaviors. AlphaPose~\cite{fang2017rmpe} is first used to extract the poses, and PoseFlow~\cite{xiu2018poseflow} is used to track the skeletons throughout each video. We then filter out the non-human related anomalies. We remove the sub-sequences in which the only anomalous object was not a person (e.g., a car) or the anomaly cannot be detected using only body poses (e.g., fire in the scene). 
See supplementary materials for the list of deleted non-HR anomalous actions.
As a result, we leave the validation set unaltered while eliminating the frames $2.32$\% of the test set. Table~\ref{tab:datasets} lists the total number of normal and abnormal frames.

\myparagraph{CUHK Avenue.} The CUHK Avenue dataset~\cite{lu13} contains 16 training videos and 21 testing videos with a total of 47 abnormal events recorded with different camera positions and angles. The HR-version \cite{morais19} is obtained by removing frames where (1) the anomalous event is non-human, (2) the person involved is occluded, or (3) the main subject cannot be detected and tracked. 

\myparagraph{ShanghaiTech Campus.} The ShanghaiTech Campus (STC) dataset~\cite{luo17} contains footage from 13 cameras around the campus with different light conditions and camera angles. It contains more than 300,000 total frames, and there are 130 abnormal events, some of which are not present in other datasets (e.g., chasing, brawling). The HR-version \cite{morais19} is obtained by removing 6 out of 107 test videos where the anomalous event is non-human. 

\subsubsection{\added{Evaluation metric}}
We score each frame in a video as mentioned in Sec.~\ref{scoring}. Then we compare it with the ground-truth labels to compute the \textit{Area Under the Curve} (AUC) score, following the previous literature in Video AD~\cite{morais19,markovitz20,lu13,acsintoae22,Georgescu21}. 
%%%%%%%%% removed liu18 \cite{liu18,morais19,markovitz20,lu13,acsintoae22,Georgescu21}.

% \input{LaTeX/tables/experiment_table}
\begin{table}[!ht]
\centering
\resizebox{\columnwidth}{!}{%
\begin{tabular}{ccccccc} 
\hline
\hline

& \multicolumn{1}{c}{\textbf{Video-based methods}}                                              &           \textbf{Params} &               \multicolumn{1}{c}{\textbf{UBnormal}}  &  \textbf{STC} &  \textbf{Avenue}  \\ 
\hline

\grayout{\textit{S}}& \grayout{\citen{Sultani} \cite{sultani18}} &   \grayout{-} & \grayout{50.3} &  \grayout{-} &  \grayout{-} \\ 
& \grayout{\citen{Georgescu}\cite{Georgescu21}}   & \grayout{$>80$M} &  \grayout{61.3} &  \grayout{-}  &  \grayout{-} \\ 
& \grayout{\citen{Bertasius}\cite{bertasius21}} & \grayout{121M} & \grayout{68.5} &  \grayout{-} &  \grayout{-} \\

\arrayrulecolor{gray}\hline\arrayrulecolor{black}\grayout{\textit{WS}} 

& \grayout{\citen{Georgescu} \cite{Georgescu21}} & \grayout{$>80$M}  & \grayout{59.3} & \grayout{82.7} & \grayout{92.3} \\

\arrayrulecolor{gray}\hline\arrayrulecolor{black}

\grayout{\textit{OCC}} %& \grayout{\citen{Hasan} \cite{hasan16}}  &  \grayout{- }& \grayout{ - } & \grayout{70.4 } & \grayout{80.0 } \\
&  \grayout{\citen{Park} \cite{park2020learning}}  &  \grayout{-} & \grayout{-} & \grayout{69.8} & \grayout{82.8}       \\
% &  \grayout{\aadded{\citen{Tang}~\cite{TANG20}}}  &  \grayout{-} & \grayout{-} & \grayout{\aadded{71.5}} & \grayout{\aadded{83.7}}       \\
&  \grayout{\citen{Tang}~\cite{TANG20}}  &  \grayout{-} & \grayout{-} & \grayout{71.5} & \grayout{83.7}      \\

%%%%%%%%%%%% removed liu18
% &  \grayout{\citen{Liu} \cite{liu18}}  &  \grayout{14M} & \grayout{-} & \grayout{72.8} & \grayout{85.1}       \\

% & \grayout{\aadded{\citen{Sun} \cite{sun24}}}  &  \grayout{\aadded{-} }& \grayout{\aadded{-} } & \grayout{\aadded{73.1} } & \grayout{\aadded{85.4} } \\
& \grayout{\citen{Sun} \cite{sun24}}  &  \grayout{-} & \grayout{-}  & \grayout{73.1}  & \grayout{85.4}  \\

&  \grayout{\citen{Chang} \cite{chang2020clustering}}  &  \grayout{-} & \grayout{-} & \grayout{73.3} & \grayout{86.0}       \\
% & \grayout{\aadded{\citen{Qiu} \cite{qiu24}}}  &  \grayout{\aadded{-} }& \grayout{\aadded{-} } & \grayout{\aadded{73.7} } & \grayout{\aadded{86.2} } \\
& \grayout{\citen{Qiu} \cite{qiu24}}  &  \grayout{-} & \grayout{- } & \grayout{73.7}  & \grayout{86.2}  \\

& \grayout{\citen{Barbalau} \cite{BARBALAU2023103656} } &  \grayout{$>80$M} &  \grayout{62.5}  & \grayout{83.8}  & \grayout{93.7}  \\

\hline
\hline

& \multicolumn{1}{c}{\textbf{Skeleton-based methods}} &  \textbf{Params} & \multicolumn{1}{c}{\textbf{HR-UBnormal}}  &  \textbf{HR-STC} &  \textbf{HR-Avenue} \\
\hline

\textit{OCC} & \citen{Morais} \cite{morais19} &  25K &  61.2  & 75.4 &  86.3 \\ 
& \citen{Markovitz} \cite{markovitz20} &  805K  &   55.2  & 74.8 &  58.1 \\
& \citen{Luo} \cite{luo19}  &  8M &  - & \underline{76.5} & \underline{87.3} \\
& Ours - \textit{radial} &  285K  & 63.4  & 75.2 &  82.2 \\
& Ours - \textit{Euclidean} & 240K &  \underline{65.2}  &  \textbf{\perfHRSTC} &  \textbf{87.8} \\
& Ours - \textit{hyperbolic}  & 240K & \textbf{65.5} &  75.6 & \underline{87.3} \\

\hline
\end{tabular}}
\caption{\added{Results on the human-related versions of the datasets UBnormal (\textit{HR-UBnormal}), ShanghaiTech Campus (\textit{HR-STC}) and CUHK Avenue (\textit{HR-Avenue)}, measured in terms of AUC score (bottom part of the table). We highlight in bold the best results and underline the second best. We report the results of video-based models on the non-HR versions of the aforementioned datasets (upper part of the table); it should be noted that such methods cannot be directly compared with skeleton-only ones, which are rather complementary, and hence are displayed in gray. The blocks split the table according to each method's framework, where \textit{S}, \textit{WS} and \textit{OCC} stand for \textit{Supervised}, \textit{Weakly-Supervised} and \textit{One Class Classification} methods, respectively.}}
\label{tab:stacked}

\end{table}

\subsection{Comparison with SoA} \label{comparison}

\added{We compare \modelname\ against relevant skeleton-based state-of-the-art techniques on the HR version of the selected three datasets. This comparison, considering skeleton-based anomaly detection algorithms, is described in Sec.~\ref{sec:skel_comp} 
and regards the bottom part of Table~\ref{tab:stacked}. 
Then, Sec.~\ref{sec:super_comp} compares skeleton-based Vs. video-based techniques, and it discusses progress on the two fronts.}

\subsubsection{Comparison with skeleton-based techniques} \label{sec:skel_comp}
\added{Table~\ref{tab:stacked} reports the results of the experiments we have
% \FG{have} 
conducted on HR-UBnormal, HR-ShanghaiTech Campus, and HR-Avenue. 
On HR-UBnormal, all three proposed volume shrinking strategies of \modelname\ introduced in Sec.~\ref{latentmethod} outperform the current best \cite{morais19}. In particular, our best variant, \modelname-\textit{hyperbolic}, outperforms \cite{morais19} by a relative improvement of 7\% (65.5 AUC Vs 61.2 AUC).
This remarks on the effectiveness of \modelname\ in the case of the challenging open set anomalies of UBnormal.}

\added{The HR-ShanghaiTech Campus and HR-Avenue rankings are tighter; still, \modelname-\textit{Euclidean} attains the best performance on both datasets, setting a new state-of-the-art on these benchmarks. Although the performance gain between \modelname-\textit{Euclidean} (77.1 and 87.8) and the current best skeleton-based method of \cite{luo19} (76.5, 87.3) is relatively small, it should be noted that, even compared to skeleton-based baselines, our model remains lightweight (cf.\ the parameter count in Table~\ref{tab:stacked})}. 
Comparing the three \modelname\ versions, the \textit{radial} approach underperforms. This might be a consequence of the more dispersed distribution of the embeddings from the center, as visible in Fig.~\ref{fig:spaces}.

\subsubsection{Relation of \modelname\ with video-based methods} \label{sec:super_comp}

\added{For completeness, the upper part of Table~\ref{tab:stacked} reports video-based techniques, tested on the complete set of videos of \textit{UBnormal}, \textit{ShanghaiTech Campus} and \textit{CUHK Avenue}. The complete datasets include human-related anomalies, as well as anomalies not relating to people (e.g., crashing cars) and only stemming from visual cues, such as fire, smoke, and fog. The general video-based anomaly detection techniques in the top part of Table~\ref{tab:stacked} have access to the spatio-temporal volume of RGB-pixel information, accounting for appearance (color, shape, object-like, etc.) and motion cues (e.g.,\ optical flow). 
Indeed, even without the ability to detect non-human events, our model achieves 65.0 AUC (cf. Table~\ref{tab:ablation_together}). In contrast, the current state-of-the-art OCC video-based model~\cite{BARBALAU2023103656} only scores an AUC of 62.5. Besides, COSKAD only uses a fraction of parameters of \cite{BARBALAU2023103656}, namely 240K Vs.\ 80M ($-99.7\%$).
On \textit{ShanghaiTech Campus} and \textit{CUHK Avenue}, our best-performing model reports an AUC score of 74.3 and 85.7, respectively, 
and only \cite{BARBALAU2023103656} when tested on \textit{CUHK Avenue} surpass \modelname's performance. 
}

\subsubsection{\addedadded{Further notes on skeleton- vs.\ video-based methods}}

\addedadded{
The overall performance of COSKAD is encouraging and employing a skeleton-based model has certain advantages over video-based competitors.
COSKAD targets human-related
anomalies.
The skeletal representation uses the motion information of the individuals appearing in a scene, exposing behavioral clues and abstracting them from potentially misleading video features, e.g., the viewing direction or the scene's illumination.
The appearance-based features of the RGB frames are not employed in COSKAD, so the algorithm neglects the anomalies which are consequences of visual anomalies such as fires (see also the limitations in Sec.~\ref{limitations}).
However, appearance-based features are complementary to the skeleton-based features of COSKAD. In fact, our models' high performance and reduced complexity suggest that skeleton-based techniques could be a valuable research direction to complement appearance-based cues. Moreover, by solely processing motion, as \modelname\ does, additional privacy and fairness guarantees are provided, as the model cannot exploit biases related to skin color, gender, or clothing style.
}

\subsection{Experimental setup} \label{exp_setup}

\myparagraph{Implementation details.} 
We train our proposed \modelname~with PyTorch Lightning
using two Nvidia P6000 GPUs for 80 epochs, with a learning rate of 0.0001 and ADAM 
%%%%%%%%% removed kingma
% \cite{kingma15} 
optimizer. The training phase required 1.5 hours, which is a fraction of the training time of \cite{morais19,markovitz20}.
We consider $V=17$ key points to represent a pose and divide each agent's motion history by adopting a sliding window procedure (each window has a length of $T=12$ frames with stride 1 so that windows overlap). 

\myparagraph{Pose normalization.}
Since the 2D positions of the joints refer to the frame, we normalize the poses to make them independent of the spatial location, following \cite{morais19}.
For all the datasets, we perform an additional normalization stage by applying robust scaling to reduce the contribution of outliers, as also done in \cite{morais19}.
\label{sec:experiments}

\section{Ablation Studies}

In this section, we report additional results and experiments that have guided us in building \modelname. This study focuses on establishing the two main components of the proposed model and compares two strategies concerning the center update and the scoring method. 
All the experiments presented in this section are performed on the \added{established} UBnormal dataset and are conducted using the best-performer Euclidean and the hyperbolic versions of \modelname. Notably, each model's variant with an STS-GCN encoder presented in this section \added{outperforms all the OCC techniques reported in Table~\ref{tab:stacked}}.

\subsection{Encoder} \label{encoderablation}
As illustrated in Sec.~\ref{encodermethod}, \modelname\ is equipped with a separable GCN-Encoder to process the input graphs. The first to propose a GCN as a kinematic encoder in the context of anomaly detection was \cite{luo19}, while previous works rely on LSTMs,
%%%%%% removed hochreiter97
% \cite{hochreiter97}, 
such as \cite{morais19}. For this reason, we compare our encoder with other established GCN architectures, such as a plain GCN~\cite{kipf16} and the ST-GCN~\cite{yan18}, which have been employed in \cite{luo19,markovitz20}.
As reported in Table \ref{tab:ablation_together}, the selected separable GCN attains the best results among competitors, showing an increase in performance of 9.8\% and 10.2\% over GCN~\cite{kipf16} and 3.6\% and 7.8\% over ST-GCN~\cite{yan18} on the Euclidean and hyperbolic models, respectively. This result confirms that separating the kinematic adjacency matrix in its spatial and temporal parts allows for improved input representations. 
Moreover, separating the adjacency matrix results in a significant reduction in parameters, yielding an encoder with only 31K parameters against 75K and 170K of \{ST-GCN, GCN\}-based encoders, respectively.

\begin{table*}[t]
\centering
\resizebox{1\textwidth}{!}{

\begin{tabular}{ccc|ccc|cc|c|cc} 
\hline
\multicolumn{3}{c|}{\textbf{Encoder}} & \multicolumn{3}{c|}{\textbf{Projector}} & \multicolumn{2}{c|}{\textbf{Center Update}} & \textbf{Hyperbolic} & \multicolumn{2}{c}{\textbf{UBnormal}} \\

\multicolumn{1}{c}{GCN\cite{kipf16}} &  \multicolumn{1}{c}{ST-GCN\cite{yan18}} & \multicolumn{1}{c|}{Sep. GCN} & \multicolumn{1}{c}{Identity} & \multicolumn{1}{c}{Linear} & \multicolumn{1}{c|}{Non-Linear} & \multicolumn{1}{c}{Static} & \multicolumn{1}{c|}{Dynamic} & & Valid.	&  Test\\ \hline \hline
  
\hline
\cmark &  &  & & & \cmark & & \cmark & & 68.9 & 59.1  \\
\cmark &  &  & & & \cmark & & \cmark & \cmark & 69.1 & 59.0 \\

 & \cmark &  & & & \cmark & & \cmark & & 68.4 & 62.6 \\
 & \cmark &  & & & \cmark & & \cmark & \cmark & 72.3 & 60.3\\

\hline
&  & \cmark  & \cmark &  &  & \cmark &  & & 72.7 & 64.5 \\
&  & \cmark  & \cmark &  &  & \cmark &  & \cmark & 72.0 & 63.6 \\

&  & \cmark  & \cmark &  &  &  & \cmark & & 71.3 & 64.7 \\
&  & \cmark  & \cmark &  &  &  & \cmark &  \cmark & 72.9  & 64.5 \\

&  & \cmark  &  & \cmark &  & \cmark &  & & 69.8 & 64.2 \\

&  & \cmark  &  & \cmark &  &  & \cmark & \cmark &  71.1  & 64.7 \\

\hline
&  & \cmark  &  &  & \cmark & \cmark &  & & 74.5 & 64.6 \\
&  & \cmark  &  &  & \cmark & \cmark &  &  \cmark & 74.0 & 63.1 \\
&  & \cmark &  & & \cmark & & \cmark & & 75.8 & 64.9 \\
&  & \textcolor{red}{\cmark} &  & & \textcolor{red}{\cmark} & & \textcolor{red}{\cmark} & \textcolor{red}{\cmark} & \textbf{76.4} &\textbf{ 65.0} \\

\hline

\end{tabular}}
\caption{Ablation on the components of the proposed method \modelname. Red checkmarks indicate the technical choices we implement in the final model. The results are attained on the UBnormal dataset.}
\label{tab:ablation_together}

\end{table*}

\subsection{Projector} \label{projectorablation}
The importance of the projector when dealing with representations and metric objectives has been studied thoroughly in the field of SSL~\cite{simclr,byol}. 
Following \cite{simclr}, we compare three different definitions for the projector: the \textit{Identity} (no projector), \textit{Linear} (two linear layers), and \textit{Non-Linear} (one block of linear layer, non-linearity, and Batch Normalization
% ~\cite{ioffe15} 
followed by another linear layer) projectors.
In Table~\ref{tab:ablation_together}, we empirically confirm the results of \cite{simclr}, showing that a non-linear projector provides a boost in performance ($+5.6\%$, $+0.54\%$ on validation and test, respectively) when compared to an Identity projector, i.e., directly using the output of the encoder. 
The Identity and Linear strategies provide similar results, showing that a non-linearity is needed to improve the representations in the latent space. 
As a further investigation, we also verify the deepness of the non-linear projector module. Specifically, we assess our model against different versions of it with a naive non-linear decoder without any block, performing only Batch Normalization, a ReLU activation and a linear layer, and a deeper projector involving two non-linear blocks followed by a linear layer. As depicted in Table~\ref{tab:projabl}, a single non-linear block considerably increases performance, especially in the hyperbolic setting ($+2\%$). Since the \modelname's modules operate in Euclidean spaces, this result is not surprising and exposes the need for a projector when dealing with non-Euclidean latent spaces. Further, using more layers results in a slight degradation in performance when using either the Euclidean or the hyperbolic latent space ($-0.2\%, -0.6\%$, respectively).

\begin{table}[t!]
\centering
\begin{tabular}{ccccc}
\hline 
\textbf{\# Blocks} & \multicolumn{2}{c}{\textbf{Euclidean}} & \multicolumn{2}{c}{\textbf{Hyperbolic}} \\
& Validation  & Test  & Validation  & Test \\ 
\hline 
\hline
0 & 71.2 & 64.1 & 72.4  & 63.7 \\ 
1 & 75.8 & 64.9 & \textbf{76.4} & \textbf{65.0} \\ 
2 & 75.7 & 64.8 & 74.01 & 64.6 \\ 
\hline   

\end{tabular}
\caption{Ablation on the depth of the Non-Linear projector proposed (cf. Sec.\ref{projectorablation}).}
\label{tab:projabl}
\end{table}

\subsection{Center Update Strategy} \label{centerablation}
Since the metric objective of our method seeks to minimize the distance between the latent representations and a point defined in the latent space, it is critical to choose the optimal rule to update the center position during training to achieve optimal mapping. \cite{ruff18} proposed to fix a point in the latent space and then perform training around it. We experiment with this method, dubbed \textit{Static}, and compare it with a novel rule to update 
the center position, dubbed \textit{Dynamic}, in Table~\ref{tab:ablation_together}. As can be seen, the \textit{Dynamic} strategy provides the best result, especially in the hyperbolic space, with an increase of $3\%$ \textit{wrt} to its \textit{Static} counterpart, while the gain in the Euclidean space ($+0.4\%$) is more marginal. We expected a similar behavior since the metric defined within the Poincaré Ball induces a distance that grows exponentially with the radius, therefore, when the learning center is left free to move, the model can take advantage of it. On the other hand, with the \textit{Static} strategy, it can be challenging for the model to minimize the distances between the embeddings and the fixed center. 

\begin{table}[t]
\centering{%
\begin{tabular}{cccc} 
\hline
\textbf{Model} & \textbf{Training } & \textbf{Inference } & \textbf{UBnormal} \\ 

\hline
\hline

\modelname  & $L_{hyp}$ &$s_{hyp}$ & 64.9 \\
\hline
& & $s_{rec}$ & 63.0 \\
\modelname -AE & $L_{hyp} + L_{rec}$ & $s_{hyp}$ & 64.1 \\
& & $s_{hyp} + s_{rec}$ & 63.4 \\

\hline

\end{tabular}}
\caption{Performance evaluation of our proposed models with COSKAD-AE. AUC score is reported for the UBnormal dataset. }
\label{tab:ablation_arch}

\end{table}

\subsection{COSKAD AutoEncoder} \label{autoencoderablation}
We also consider a multi-task learning objective for \modelname\ coupling the original objective illustrated in Eq.~\ref{eq:svdd} with a reconstruction error; to do this, we included an additional module that acts as a decoder, allowing \modelname\ to reconstruct the original poses. Altogether, this makes our proposed model a GCN AutoEncoder, dubbed \modelname-AE, and the new module has been obtained by reversing the GCN-Encoder.
Table \ref{tab:ablation_arch} compares the performance of \modelname-AE, trained with both reconstruction and hypersphere loss, against our best model. 
Three methods for calculating the anomaly score are possible: either the reconstruction or hypersphere loss can be employed alone (named $s_{rec}$ and $s_{hyp}$, respectively in Table ~\ref{tab:ablation_arch}), or the two scores can be combined ($s_{rec} + s_{hyp}$). The best performances are achieved by the only hypersphere score. This is probably due to the Separable GCN-Encoder that, separately learning the trajectories of single joints and unified poses for each timeframe, exposes improved generalization that affects $s_{rec}$ and provides better reconstructions.

\label{sec:ablation}

\begin{figure}[!t]
    \centering
    \begin{subfigure}{0.3\textwidth}
         \centering
         \includegraphics[width=\textwidth]{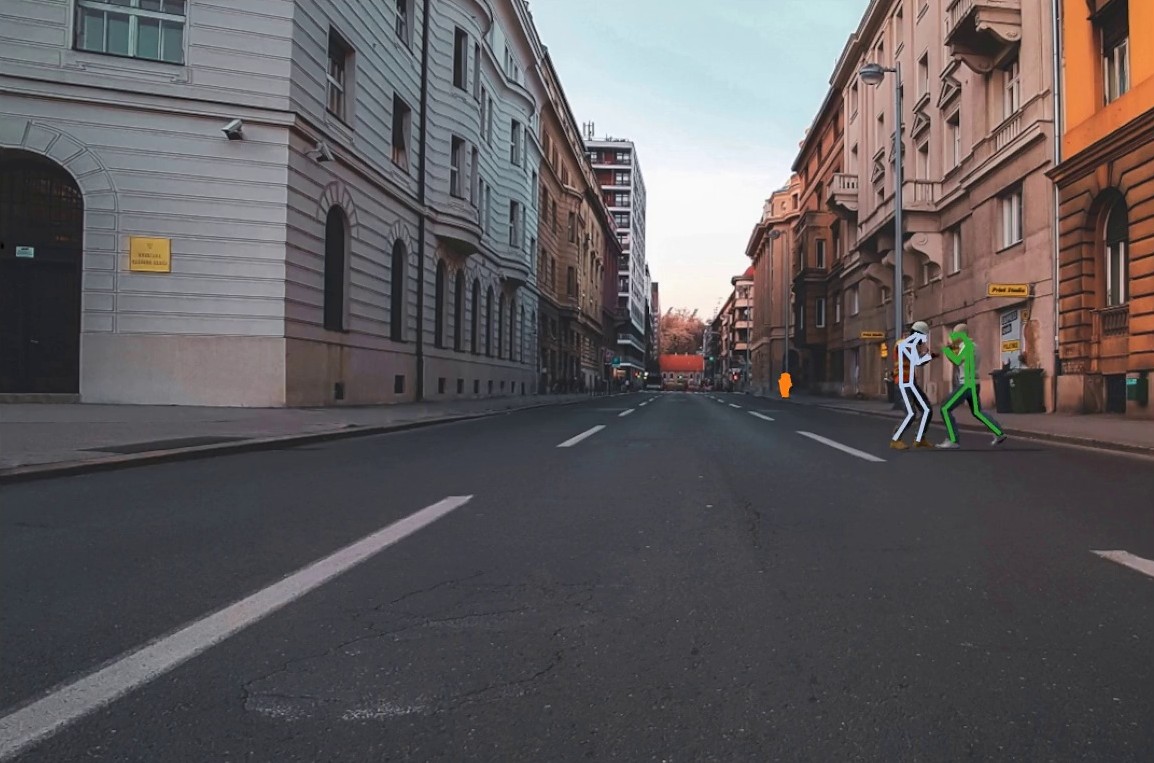}
         \caption{}\label{fig:1a}
    
    \end{subfigure}%
    \hfill
    \begin{subfigure}{0.3\textwidth}
         \centering
         \includegraphics[width=\textwidth]{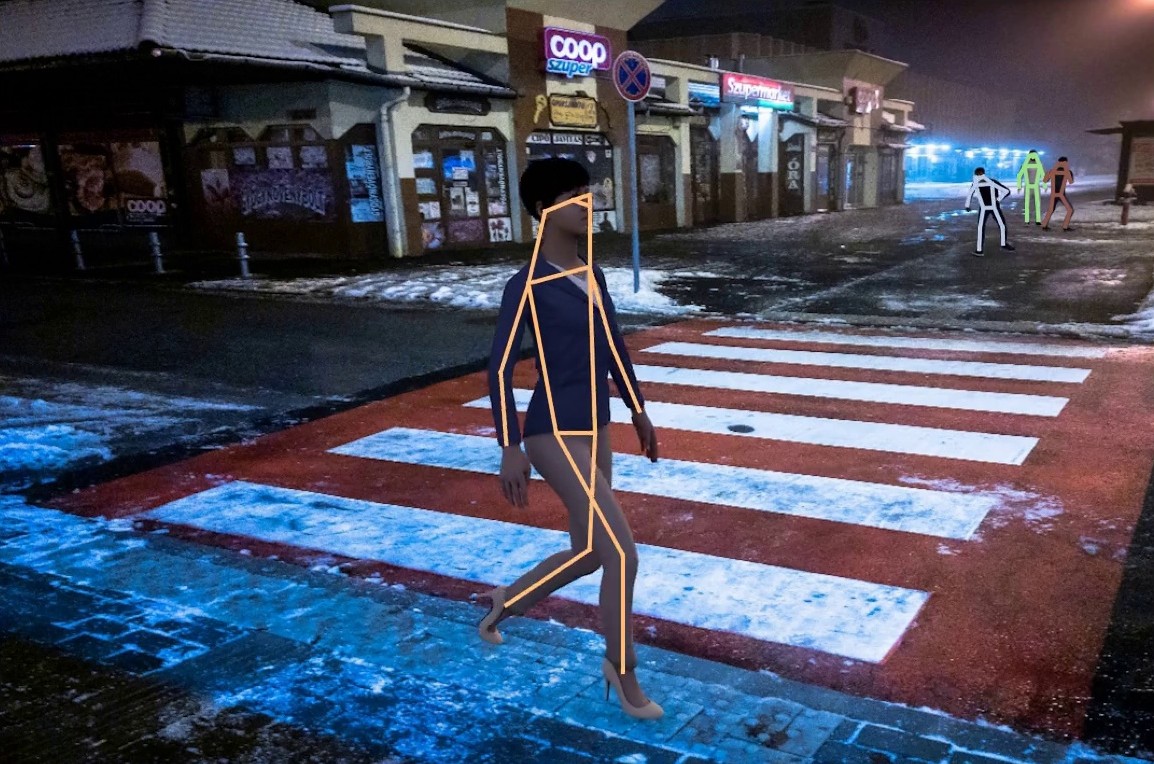}
         \caption{}\label{fig:1b}
    \end{subfigure}%
    \hfill
    \begin{subfigure}{0.3\textwidth}
         \centering
         \includegraphics[width=\textwidth]{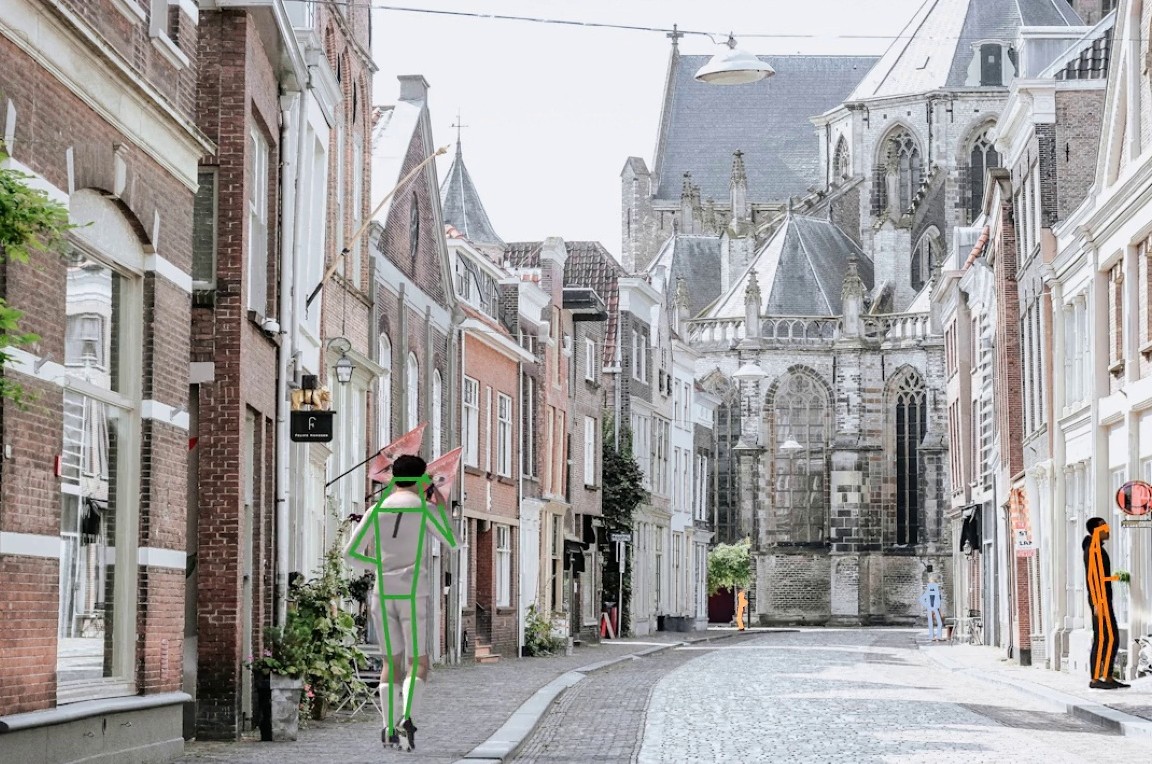}
         \caption{}\label{fig:1c}
    \end{subfigure}%

\caption{\addedadded{Examples of extracted poses in \hrub. The poses are correctly detected even in challenging conditions, e.g., different scales or unusual poses. See section \textit{Sample of misestimated human poses} for discussion.}}
\label{fig:1}
\end{figure}

\begin{figure}[!t]
    \centering
    \begin{subfigure}{0.3\textwidth}
         \centering
         \subfloat[]{\label{fig:2a}\includegraphics[width=\textwidth]{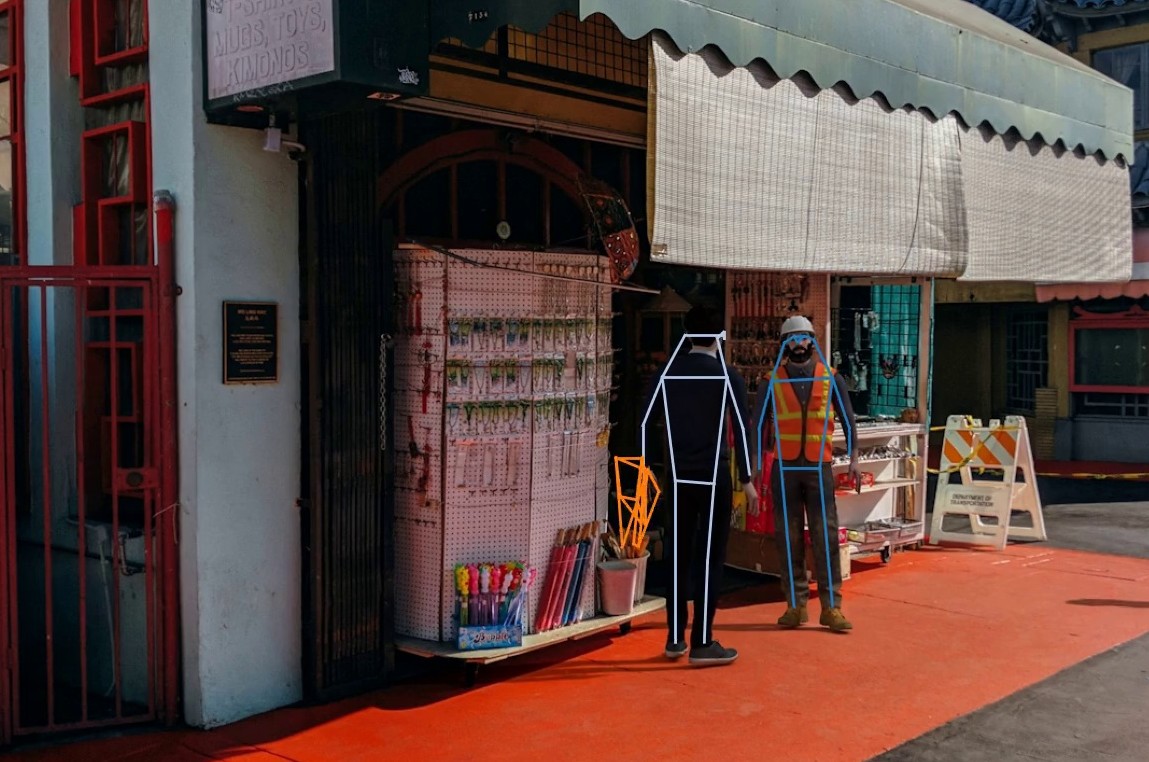}}
    \end{subfigure}%
    \hfill
    \begin{subfigure}{0.3\textwidth}
         \centering
         \subfloat[]{\label{fig:2b}\includegraphics[width=\textwidth]{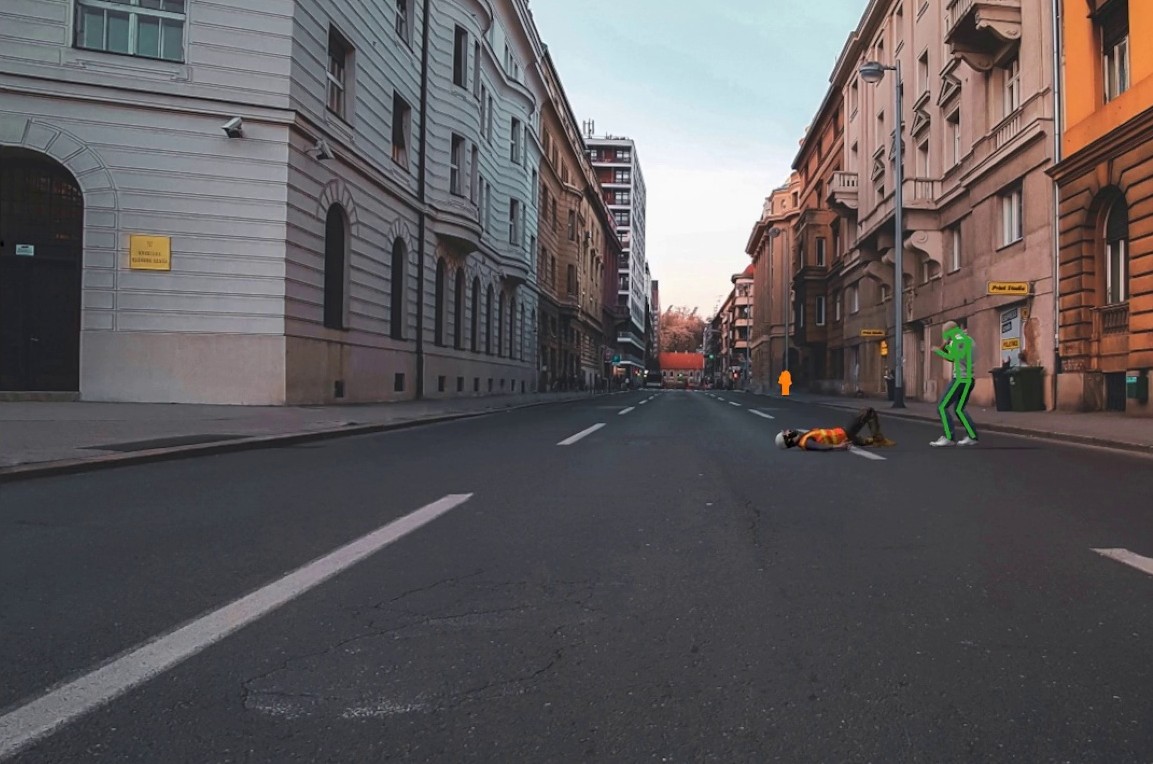}}
    \end{subfigure}%
    \hfill
    \begin{subfigure}{0.3\textwidth}
         \centering
         \subfloat[]{\label{fig:2c}\includegraphics[width=\textwidth]{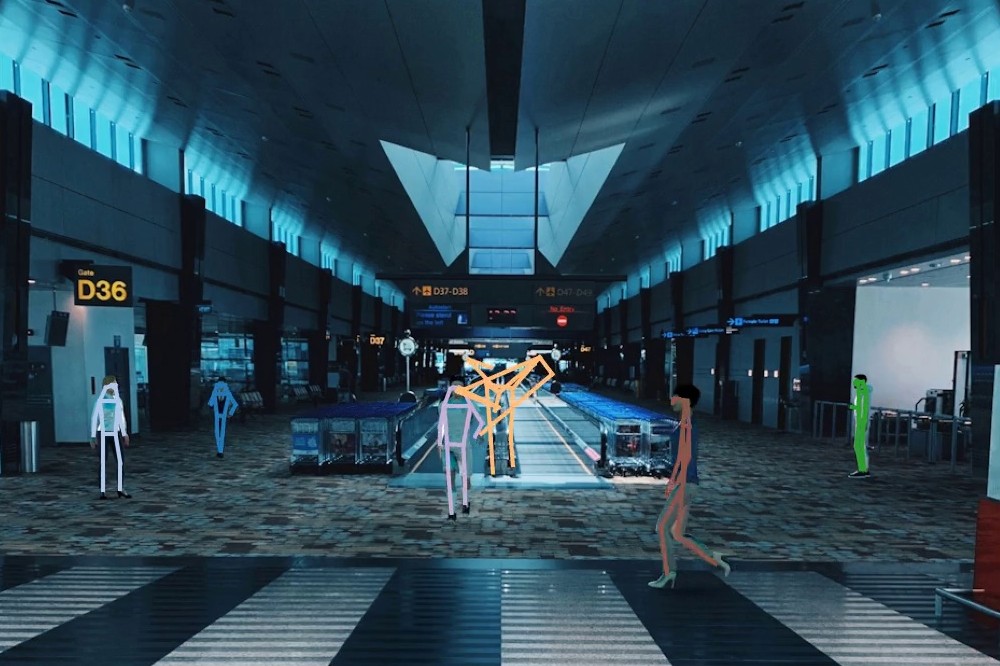}}
    \end{subfigure}%

\caption{\addedadded{Examples of misestimations of the pose extractor in \hrub. Fig. \ref{fig:2a} shows a pose that is not present in the scene, Fig.~\ref{fig:2b} is an example of a pose that is not detected. Fig.~\ref{fig:2c} is an example of a noisy pose estimation due to the scale of the subject and its partial occlusion. See section \textit{Sample of misestimated human poses} for discussion.}}
\label{fig:2}
\end{figure}

\section{\addedadded{Limitations}}\label{limitations}

\addedadded{In Sec.~\ref{sec:experiments} we prove the significance of our proposed \modelname~in the context of skeleton-based Video Anomaly Detection. However, despite the promising performance revealed, specific challenges persist, primarily related to misestimated skeletal poses occurring in the datasets and those due to the model shortcomings. These issues are commented in the following Sec.~\ref{sec:misestimated_poses} and Sec.~\ref{sec:model_shortcomings}, respectively.}

\subsection{\addedadded{Samples of misestimated human poses}}\label{sec:misestimated_poses}

\addedadded{Despite being extracted from a synthetic dataset, \hrub~poses have been obtained using a system that yields effective outcomes. In Fig.~\ref{fig:1}, we have included 3 frames from the proposed \hrub~ in which Alphapose correctly extracts poses of agents in complex positions (Fig.~\ref{fig:1a}) or at very different scales (Fig.~\ref{fig:1b},~\ref{fig:1c}). Nevertheless, Alphapose is not error-free, and in this section, we discuss some limitations of our dataset. Fig.~\ref{fig:2} illustrates some misestimations performed by Alphapose: in Fig.~\ref{fig:2a}, a pose is incorrectly detected from the background near two perfectly estimated people, Fig.~\ref{fig:2b} shows a lying agent whose pose is not detected by Alphapose, probably due to the supine pose assumed, and in Fig.~\ref{fig:2c} it can be seen that the system detects a noisy pose due to the subject's scale and partial occlusion. While we have tried to mitigate this problem by removing many noisy poses, some failure cases are challenging to recognize because they may last only one or a few frames.}

\begin{figure}[t]
    \centering
    \begin{subfigure}{0.45\textwidth}
         \centering
         \subfloat[]{\label{fig:3a}\includegraphics[width=\textwidth]{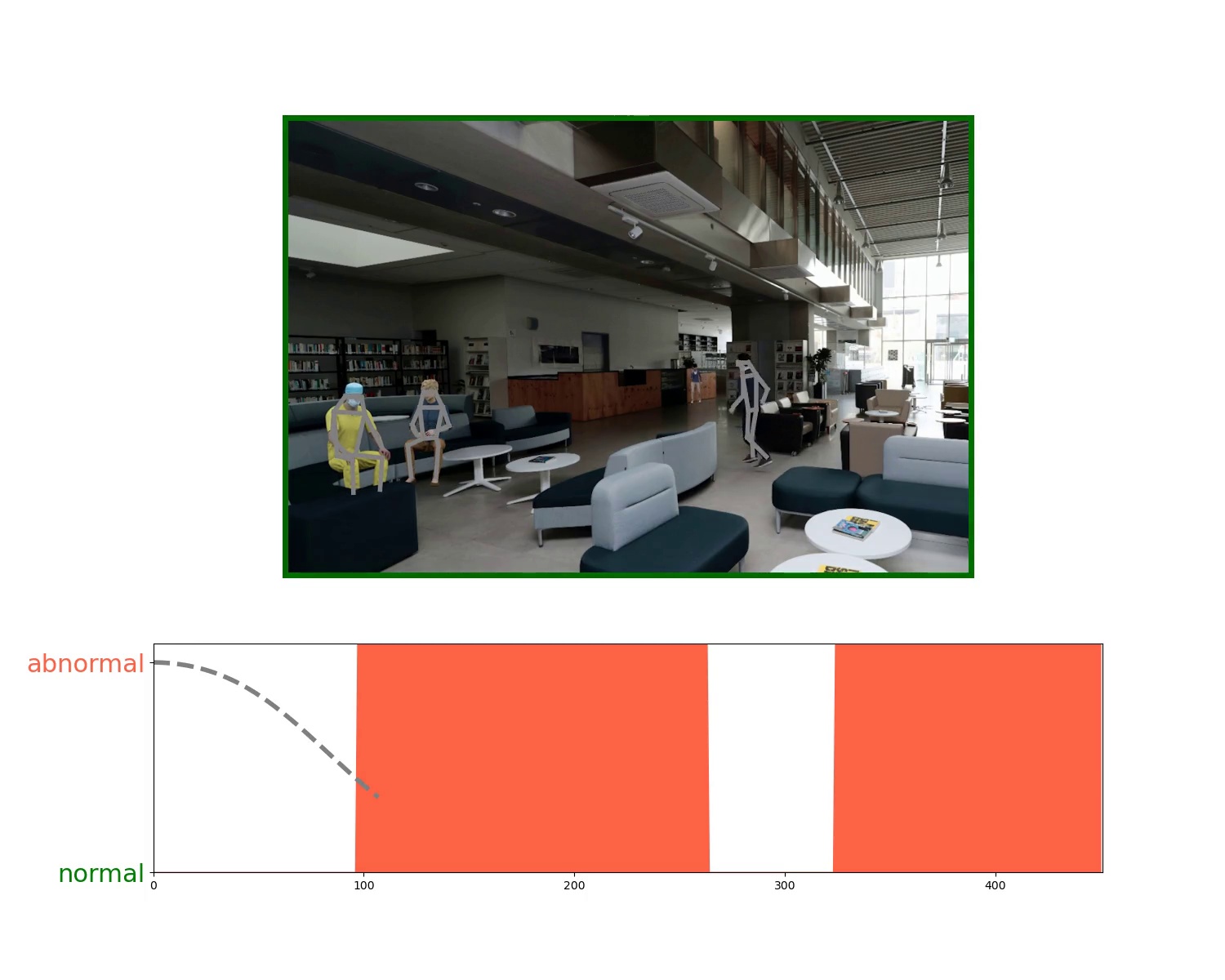}}
    \end{subfigure}%
    \hfill
    \begin{subfigure}{0.45\textwidth}
         \centering
         \subfloat[]{\label{fig:3b}\includegraphics[width=\textwidth]{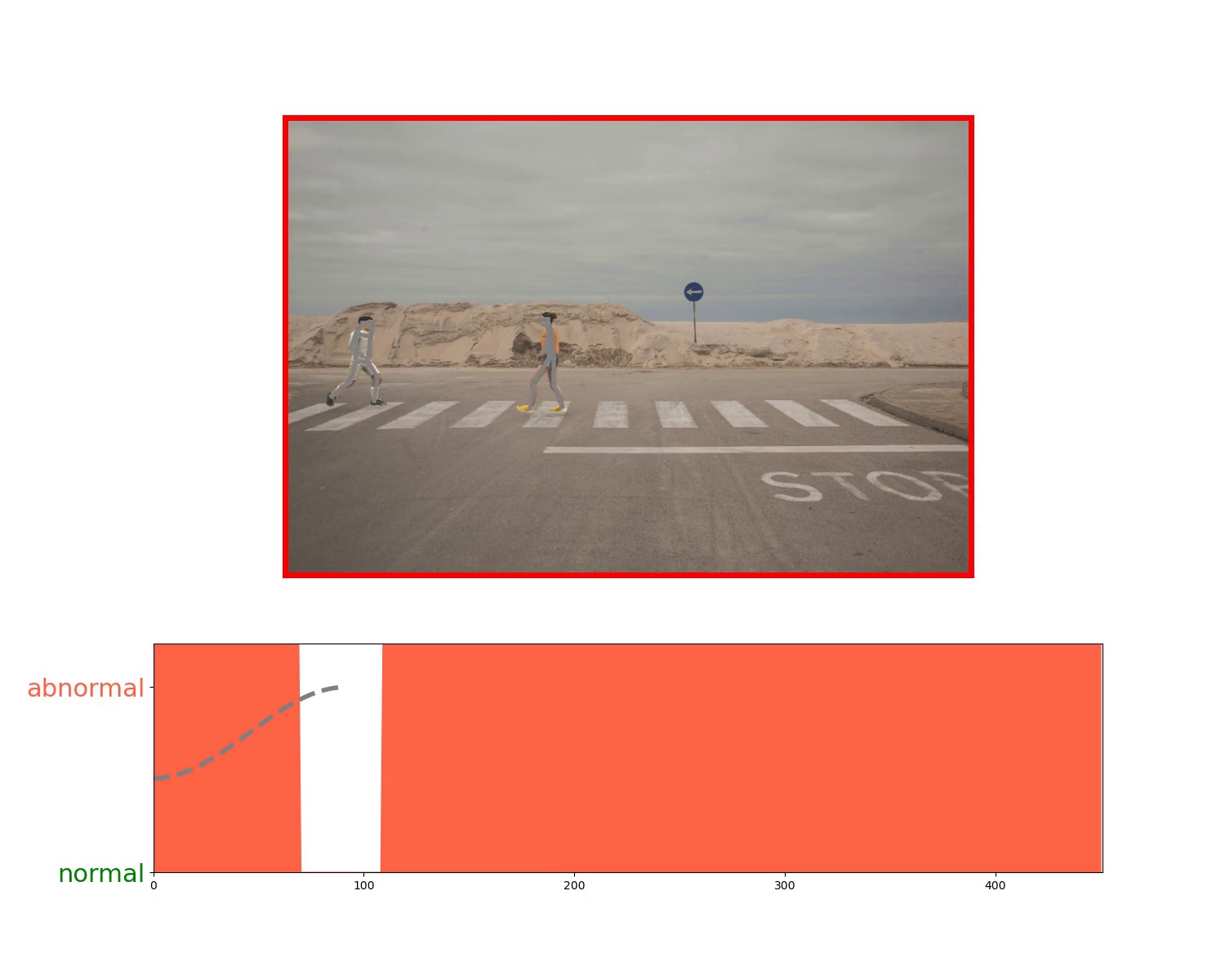}}
    \end{subfigure}%

\caption{\addedadded{Examples of failure cases from the test set of \hrub, and the extracted score of the frame assigned by our proposed \modelname. \textit{(left)} the standing subject is dancing, but it is not detected as anomalous (\textit{false negative}). \textit{(right)} people depicted in the scene are walking, but the model predicts them as anomalous (\textit{false positive}). See section \textit{Sample anomaly detection and failure cases} for a broader discussion.}}
\label{fig:3}
\end{figure}

\subsection{\addedadded{Sample of \modelname\ shortcomings}} \label{sec:model_shortcomings}

\addedadded{As mentioned in the section above, we have released several videos demonstrating our method's effectiveness and efficiency in detecting anomalies in situations of varying complexity. However, our system has some faults, and we will examine a few of them here. Fig.~\ref{fig:3} depicts two examples of our method's failure, one showing a false positive (Fig.~\ref{fig:3a}) and the other a false negative (Fig.~\ref{fig:3b}). 
In the first, the person standing is dancing, and the system does not consider it abnormal, as evidenced by the declining slope in the anomaly score graph below. On the other hand, we show in Fig.~\ref{fig:3b} how \modelname~ struggles to be accurate when normal and abnormal actions are interspersed. One of the agents in this scene starts running, stops, and then resumes running. The graph demonstrates that the anomaly score is still high even during the clip's regular portion.}

\label{sec:limitations}

\section{Conclusion}
We have proposed a novel Skeleton-based anomaly detection method based on the minimization of latent vectors to a center, exploiting the properties of three different manifolds: Euclidean, hyperbolic, and spherical. We defined the minimization metrics and scoring and investigated the alterations in space induced by \aadded{these} manifolds. By leveraging an \added{STS-}GCN encoder and coupling it with a loss that matches the OCC objective, \modelname\ outperforms SoA models on established \added{human-related} benchmarks. On the most recent and challenging HR-UBnormal, all three versions of the proposed \modelname\ reach SoA performance, which shows the representational power of our approach. 
\label{sec:conclusion}

\bibliography{biblio}

\paragraph{Alessandro Flaborea}
 completed the Ph.D. at the Sapienza University of Rome. He received his degree in Computer Science from the University of Udine, Italy, and his master's degree in Data Science from the Sapienza University of Rome.  
His research interests include video anomaly detection, procedural learning, and hyperbolic neural networks.

\paragraph{Guido Maria D'Amely di Melendugno}
\aadded{finished his} Ph.D. in Computer Science at the Sapienza University of Rome. After obtaining his master's degree in mathematics from the Sapienza University of Rome, he focused on computer vision. His research interests lie in multimodal CV, pose forecasting, video anomaly detection, and procedural learning.

\paragraph{Stefano D'Arrigo}
is a Ph.D. student in Artificial Intelligence at the Sapienza University of Rome. He received a BSc in Computer Science at the University of Catania, Italy, and an MSc in Data Science at the Sapienza University of Rome. His research interests include pose estimation, video anomaly detection, and geometric deep learning.

\paragraph{Marco Aurelio Sterpa}
is an MSc student at the Sapienza University of Rome. He received his bachelor's degree in Computer Science from the Sapienza University of Rome and is currently a second-year master's degree student in Data Science.
His research interests include time series and video anomaly detection.

\paragraph{Alessio Sampieri} is a Ph.D. student in Data Science at the Sapienza University of Rome. He received his degree in Statistics for Management and master's degree in Data Science from the Sapienza University of Rome. His research interests include human motion and its forecasting, video analysis and understanding, and geometric deep learning.

\paragraph{Fabio Galasso}
heads the Perception and Intelligence Lab (PINLab) at the Dept.\ of Computer Science, Sapienza University of Rome (Italy). 

His research interests include distributed and multi-agent intelligent systems, perception (detection, recognition, re-identification, forecasting) and general intelligence (reasoning, meta-learning, domain adaptation), within sustainable (low-power-consumption and constrained-computational-resource sensors and devices) and interpretable frameworks.

\end{document}